%% file: eccv2022submissionCR.tex
\documentclass[runningheads]{llncs}
\usepackage{graphicx}
\usepackage{amsmath,amssymb}
\usepackage{booktabs}
\usepackage{comment}
\usepackage{tikz}

\usepackage{xcolor}
\usepackage{caption}
\usepackage{colortbl}
\usepackage{multirow}
\usepackage{mmstyle}
\usepackage[pagebackref,breaklinks,colorlinks]{hyperref}
\usepackage{bbding}

\usepackage[capitalize]{cleveref}
\crefname{section}{Sec.}{Secs.}
\Crefname{section}{Section}{Sections}
\Crefname{table}{Table}{Tables}
\crefname{table}{Tab.}{Tabs.}

\input{macros.tex}

\usepackage[accsupp]{axessibility}  

\begin{document}
	
	\pagestyle{headings}
	\mainmatter
	\def\ECCVSubNumber{1947}  
	
	\title{BungeeNeRF: Progressive Neural Radiance Field for Extreme Multi-scale Scene Rendering}
	
	
	
	
	
	\titlerunning{BungeeNeRF}
	
	\author{Yuanbo Xiangli\inst{1}$^*$ \and
		Linning Xu\inst{1}$^*$ \and
		Xingang Pan\inst{2} \and
		Nanxuan Zhao\inst{1,3} \and \\
		Anyi Rao\inst{1} \and
		Christian Theobalt\inst{2} \and
		Bo Dai\inst{4}\Envelope \and
		Dahua Lin\inst{1,4}}
	\authorrunning{Y. Xiangli et al.}
	%
	\institute{
		$^{1}$The Chinese University of Hong Kong \quad
		$^{2}$Max Planck Institute for Informatics \quad
		$^{3}$University of Bath \quad
		$^{4}$Shanghai Artificial Intelligence Laboratory \\
		{\tt\small \{xy019,xl020,ra018,dhlin\}@ie.cuhk.edu.hk}~
		{\tt\small \{xpan,theobalt\}@mpi-inf.mpg.de}~ \\
		{\tt\small nanxuanzhao@gmail.com}~
		{\tt\small daibo@pjlab.org.cn}
	}
	
	{
		\renewcommand{\thefootnote}%
		{\fnsymbol{footnote}}
		\footnotetext[1]{Equal contribution.}
	}
	
	
	\maketitle
	
	\input{sections/abstract}
	\input{sections/intro}

	\input{sections/related}

	\input{sections/method}
	\input{sections/experiment}
	\input{sections/ablation}
	\input{sections/conclusion}

	%
	%
	
	\newpage
	\bibliographystyle{splncs04}
	\bibliography{egbib}
\end{document}

%% file: macros.tex
\definecolor{yellow}{rgb}{1,1, 0.6}
\definecolor{lightyellow}{rgb}{1,1, 0.8}
\definecolor{orange}{rgb}{1, 0.8, 0.6}
\definecolor{coral}{RGB}{246,131,65}
\definecolor{pinkred}{rgb}{1, 0.6, 0.6}
\definecolor{hotpink}{RGB}{238,64,195}

\definecolor{amber}{rgb}{1.0, 0.49, 0.0}

\definecolor{lavender}{RGB}{207,226,243}
\definecolor{gainsboro}{RGB}{208,224,227}
\definecolor{gainsboro2}{RGB}{217,234,211}
\definecolor{blanchedalmond}{RGB}{252,229,205}

\newcommand{\modelname}{BungeeNeRF\xspace}

\newcommand{\coverage}{spatial coverage\xspace}

\newcommand{\altitude}{distance\xspace}

\newcommand{\altitudes}{distances\xspace}

\newcommand{\copyrightgoogle}{\href{https://www.google.com/help/terms_maps/}{\textcopyright 2022 Google}\xspace}

\usepackage{wrapfig}
\usepackage{soul}

%% file: sections/abstract.tex
\begin{abstract}
Neural radiance fields (NeRF) has achieved outstanding performance in modeling 3D objects and controlled scenes, usually under a single scale.
In this work, we focus on multi-scale cases where large changes in imagery are observed at drastically different scales. This scenario vastly exists in real-world 3D environments, such as city scenes, with views ranging from satellite level that captures the overview of a city, to ground level imagery showing complex details of an architecture; and can also be commonly identified in landscape and delicate minecraft 3D models.
The wide span of viewing positions within these scenes yields multi-scale renderings with very different levels of detail, 
which poses great challenges to neural radiance field and biases it towards compromised results.
To address these issues, we introduce \modelname, a progressive neural radiance field that achieves level-of-detail rendering across drastically varied scales.
Starting from fitting distant views with a shallow base block, as training progresses, new blocks are appended to accommodate the emerging details in the increasingly closer views.
The strategy progressively activates high-frequency channels in NeRF's positional encoding inputs and successively unfolds more complex details as the training proceeds.
We demonstrate the superiority of \modelname in modeling diverse multi-scale scenes with drastically varying views on multiple data sources (city models, synthetic, and drone captured data) and its support for high-quality rendering in different levels of detail.
\end{abstract}

%% file: sections/intro.tex
\section{Introduction}
\label{sec:intro}

Neural volumetric representations~\cite{Mescheder2019OccupancyNL,Park2019DeepSDFLC,Lombardi2019NeuralV,mildenhall2020nerf,Liu2020NeuralSV,tewari2021advances} have demonstrated remarkable capability in representing 3D objects and scenes from images. 
Neural radiance fields (NeRF)~\cite{mildenhall2020nerf} encodes a 3D scene with a continuous volumetric function parameterized by a multilayer perceptron (MLP), and maps a 5D coordinate (position and viewing direction) to the corresponding color and volume density in a scene. While NeRF has been shown effective in controlled environments with a bounded and single-scale setting, the ``single-scale" assumption can easily be violated in real-world scenarios. Despite early attempts on representing multi-scale 3D scenes have been made in~\cite{barron2021mip,barron2022mip,lindell2021bacon}, it remains unclear how well a neural radiance field can handle scenarios at drastically varied scales.

\begin{figure}[t!]
	\centering
	\includegraphics[width=\textwidth]{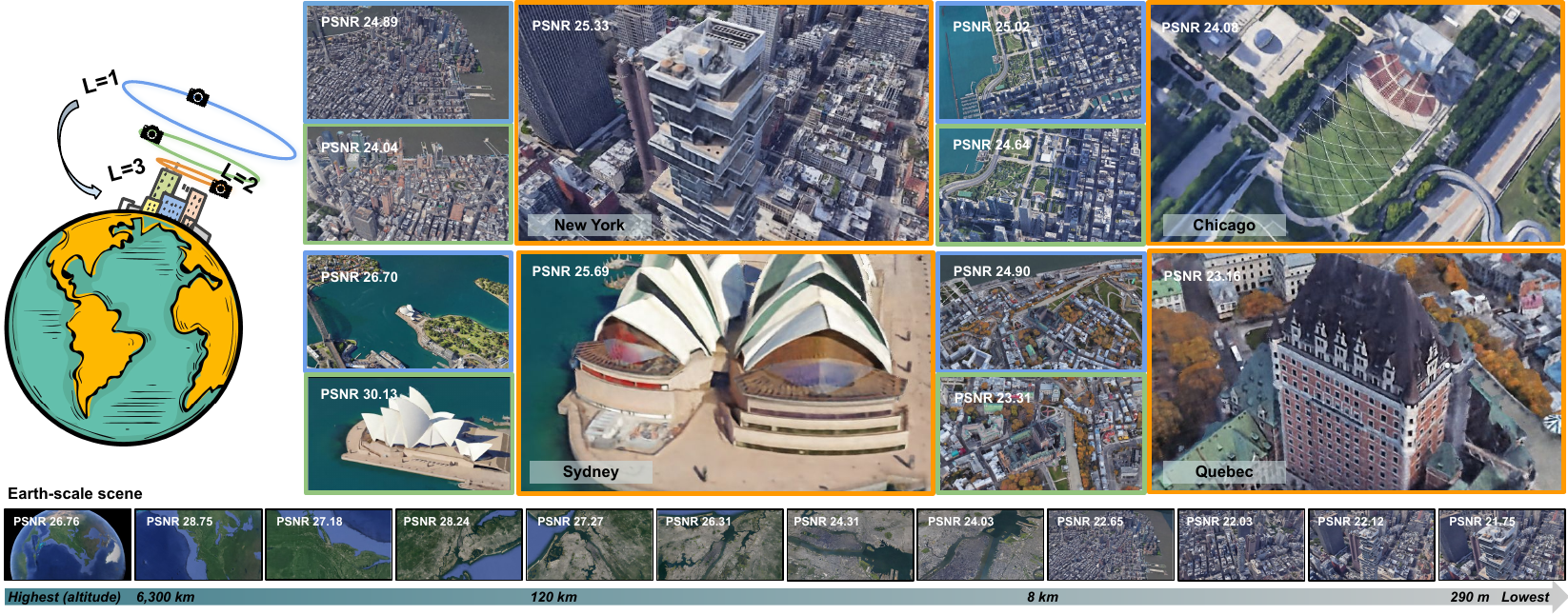}
	\caption{\small \modelname is capable of packing extreme multi-scale city scenes into a unified model, which preserves high-quality details across scales varying from satellite-level to ground-level. \emph{Top:} We use the edge colors to denote three scales from the most remote to the closest, with PSNR values displayed at the top-left corner of each rendered image. \emph{Bottom:} \modelname can even accommodate variations at \emph{earth-scale}. 
		(src: \emph{New York}, \emph{Chicago}, \emph{Sydney}, and \emph{Quebec} scenes \copyrightgoogle).}
	\label{fig:teaser}
\end{figure}

In this work, we are interested in offering high-quality renderings of scenes under such extreme scale variations.
One typical scenario is our living city, which is essentially large- and multi-scale with adequate variety in components.
A direct observation on capturing city scenes is that, the large span of allowed viewing positions within the scene can induce significant visual changes: as the camera ascends, the imagery of ground objects exhibit less geometric detail and lower texture resolution; meanwhile, new objects from peripheral regions are getting streamingly included into the view with growing \coverage, 
as shown in Fig.~\ref{fig:teaser}.
As the result, the variation in levels of detail and linear field of view raises a tension among scales when learning to model such scenes with neural representations.
It poses great challenges on NeRF and its derivatives which treat all pixel signals indiscriminately regardless of their scales.
With limited model capacity, the rendered close views always have blurry textures and shapes, and remote views tend to be incomplete with artifacts in peripheral scene areas. Other than city scenes, such challenges also commonly exist in synthetic scenes, for example in minecraft model and delicate 3D objects.

Targeting the above challenges,
we propose \modelname,
a progressive neural radiance field that enables level-of-detail rendering across the drastically varied scales.
Starting from the most remote scale, we gradually expand the training set by one closer scale at each training stage, and synchronously grow the model with multiple output heads for different level-of-detail renderings.
In this way, \modelname robustly learns a hierarchy of radiance representations across all scales of the scene.
To facilitate the learning, two special designs are introduced:
1) 
\emph{Growing model with residual block structure}:
instead of naively deepening the MLP network, we grow the model by appending an additional block per training stage. 
Each block is re-exposed to the input position encoding via a skip layer, and has its own output head that predicts the color and density residuals between successive stages, which encourages the block to focus on the emerging details in closer views and reuse these high-frequency components in the input position encoding;
2) 
\emph{Inclusive multi-level data supervision}: 
the output head of each block is supervised by the union of images from the most remote scale up to its corresponding scale. 
In other words, the last block receives supervision from all training images while the earliest block is only exposed to the images of the coarsest scale. 
With such design, each block module is able to fully utilize its capacity to model the increasingly complex details in closer views with the input Fourier position encoding features, and guarantees a consistent rendering quality between scales.

Extensive experiments show that, in the presence of multi-scale scenes with large-scale changes, \modelname is able to construct more complete remote views and brings out significantly more details in close views, as shown in Fig.~\ref{fig:teaser},
whereas baseline NeRF/Mip-NeRF trained under vanilla scheme constantly fail.
Specifically,
our model effectively preserves scene features learnt on remote views, 
and actively access higher frequency Fourier features in the positional encoding to construct finer details for close views.
Furthermore, \modelname allows views to be rendered by different output heads from shallow to deep blocks, providing additional flexibility of viewing results in a level-of-detail manner.

%% file: sections/related.tex
\section{Related Work}
\label{sec:related}

\indent \textbf{NeRF and beyond.}
NeRF has inspired many subsequent works that
extend its continuous neural volumetric representation for more practical scenarios beyond simple static scenes, including 
unbounded scenes~\cite{zhang2020nerf++,barron2022mip}, dynamic scenes~\cite{li2021neural,xian2021space}, nonrigidly deforming objects~\cite{pumarola2021d,park2021nerfies,park2021hypernerf,tretschk2021non}, phototourism settings with changing illumination and occluders~\cite{martin2021nerf,rematas2021urban}, etc.
In this work, we deal with a more extreme scenario in terms of the spatial span, and
aim to bring neural radiance fields with an unprecedented multi-scale capability, such as that in a city scenario where images can be captured from satellite-level all the way down to the ground.

While this extreme multi-scale characteristic is often compounded with the ``large scale" characteristic, such as delivering an entire city model containing hundreds of city blocks, in this project, we separate the large scenes in single-scale cases~\cite{tancik2022block,turki2021mega} from our study focus.
Orthogonal to these works which focus on dealing with the large horizontal span of urban data with proposed division and blending solutions using multiple NeRF models, we try to enlarge NeRF's capability in representing multi-scale scenes under a new training paradigm.
%

\noindent \textbf{Multi-scale Representations in 3D scenes.}
The adoption of position encoding in NeRF enables the multilayer perceptron (MLP) to learn high-frequency functions from the low-dimensional coordinate input~\cite{tancik2020fourier}.
A line of works have been proposed to tackle the multi-scale issue or coarse-to-fine learning problem by adjusting this position encoding.
In particular, Mip-NeRF~\cite{barron2021mip} uses integrated positional encoding (IPE) that replaces NeRF's point-casting with cone-casting, 
which allows the model to explicitly reason about 3D volumes. 
\cite{park2021hypernerf,lin2021barf,park2021nerfies} alternatively adopt windowed positional encoding to aid learning dynamic shapes via a coarse-to-fine training.
BACON~\cite{lindell2021bacon} differs from these works by designing multi-scale architecture that achieves mult-iscale outputs via Multiplicative Filter Networks~\cite{fathony2020multiplicative}.
While these approaches have shown multi-scale properties on common objects and small scenes, their generalization to drastic scale changes~(\eg city scenes) remain unexplored, and often caught unstable training in practice.

In image processing, many techniques relies on image pyramid~\cite{Simoncelli1995TheSP} to represent images at multiple resolutions. A related concept to multi-scale representation in computer graphics is \emph{level-of-detail} (LOD)~\cite{clark1976hierarchical,luebke2003level}, which refers to the complexity of a 3D model representation, in terms of metrics such as geometric detail and texture resolution.
The hierarchical representation learnt by \modelname can be seen as an analogue to this concept, where features learnt at different blocks represent geometries and textures of different complexity.
\cite{takikawa2021neural} 
proposed to represent implicit surfaces
using an octree-based feature volume which adaptively
fits shapes with multiple discrete LODs.
\cite{martel2021acorn} used a multi-scale
block-coordinate decomposition approach that is
optimized during training.
Recently, \cite{muller2022instant} propose a multiresolution structure with hash tables storing trainable feature vectors that achieves a combined speedup of several orders of magnitude.
These approaches work under the assumption of bounded scenes, where the covered 3D volume can be divided into multi-granularity grids or cells, whose vertices store scene features at different scales. 
Unlike these feature-based multi-scale representations, we take an alternative approach to represent multi-scale information with a dynamic model grown in blocks, which allows representing scenes of arbitrary size, instead of relying on pre-learnt features on vertices within a bounded scene.

%% file: sections/method.tex
\section{\modelname}
\label{sec:method}
In this paper, we aim at representing extreme multi-scale application scenarios with a progressive neural radiance field, where scenes are captured by cameras of very different \altitudes towards the target. 
For intuitive demonstration, our methodology is majorly elaborated in the context of city scenes. Fig.~\ref{fig:teaser} illustrates that,
the drastic span in camera \altitude brings extreme multi-scale characteristic in renderings, induced by
large-scale change in levels of detail and linear field of view.

In the following sections, Sec.~\ref{sec:preliminary} gives the necessary background for NeRF and Mip-NeRF. 
Sec.~\ref{subsec:observation} discusses
the challenges of
representing scenes under the drastic multi-scale condition with neural radiance field.
Our proposed progressive network growing and training scheme is elaborated in Sec.~\ref{subsec:method}.

\begin{figure*}[t!]
	\centering
	\includegraphics[width=\linewidth]{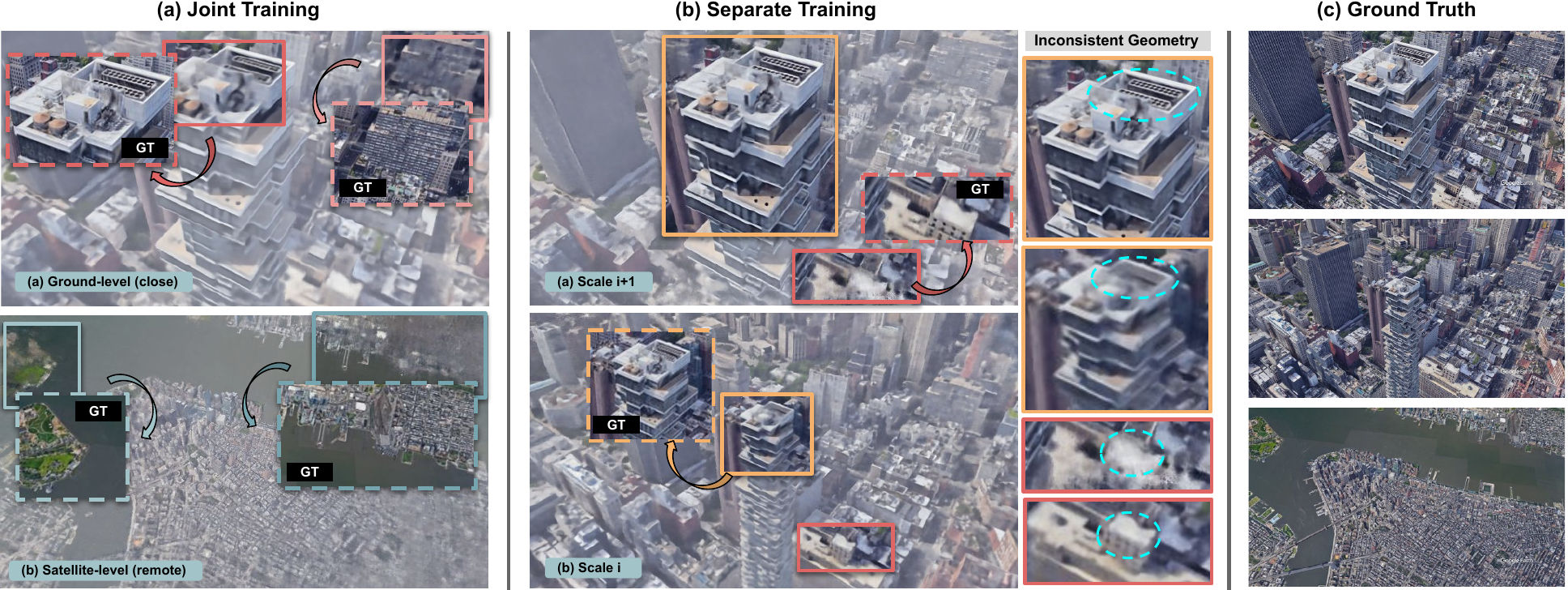}
	\caption{\small Observed artifacts on modeling multi-scale scenes with neural radiance fields~\cite{barron2021mip}: (a) jointly train on all scales and (b) separately on each scale. Artifacts on rendered images are highlighted by solid boxes, with ground truth patches (GT) displayed on the side. Joint training on all scales results in blurry texture in close views and incomplete geometry in remote views; while separate training on each scale yields inconsistent renderings between successive scales, where an additional fusion step is generally required.}
	\label{fig:observation_joint_separate}
\end{figure*}

\subsection{Preliminaries on NeRF and Mip-NeRF}
\label{sec:preliminary}
NeRF \cite{mildenhall2020nerf} parameterizes the volumetric density and color as a function of input coordinates, using the weights of a MLP.
For each pixel on the image,
a ray $\mathbf{r}(t)$ is emitted from the camera's center of projection and passes through the pixel. 
For any query point $\mathbf{r}(t_{k})$ on the ray, the MLP takes in its Fourier transformed features,~\ie position encoding (PE), and outputs the per-point density and color.
Mip-NeRF~\cite{barron2021mip} treats rays as cones and extends NeRF's point-based input to volumetric frustums, which remedies NeRF's aliasing issue when rendering at varying resolution. For each interval $T_k = [t_k,t_{k+1})$ along the ray, the conical frustum is represented by its mean and covariance  $(\mu,\mathbf{\Sigma}) = \mathbf{r}(T_k)$ which is further transformed to Fourier features with integrated positional encoding (IPE):
\begin{equation*}
\resizebox{0.65\textwidth}{!}{
	$
	\gamma(\mathbf{\mu},\mathbf{\Sigma}) =
	\left \{ \left[
	\begin{aligned}\sin (2^{m}\mathbf{\mu}) \exp(-2^{2m-1} \diag (\mathbf{\Sigma})) \\
		\cos (2^{m}\mathbf{\mu}) \exp(-2^{2m-1} \diag (\mathbf{\Sigma})) 
	\end{aligned}
	\right]\right \}^{M-1}_{m=0},
	$
}
	\label{eq:ipe}
\end{equation*}
based on a Gaussian approximation over the conical frustum. 
The MLP is then optimized abiding by the classical volume rendering, where
the estimated densities and colors for all the sampled points $\mathbf{r}(t_{k})$ are used to approximate the volume rendering integral using numerical quadrature \cite{max1995optical}:
\begin{equation*}
\resizebox{\textwidth}{!}{
	$
		\mathbf{C}(\mathbf{r} ; \mathbf{t})=\sum_{k} T_{k}\left(1-\exp \left(-\tau_{k}\left(t_{k+1}-t_{k}\right)\right)\right) \mathbf{c}_{k}, \quad T_{k}=\exp \left(-\sum_{k^{\prime}<k} \tau_{k^{\prime}}\left(t_{k^{\prime}+1}-t_{k^{\prime}}\right)\right),
	$
}
\label{eq:integral}
\end{equation*}
where $\mathbf{C}(\mathbf{r}; \mathbf{t})$ is the final predicted color of the pixel. 
The final loss is the total squared error between the rendered and ground truth colors for the collection of sampled rays within a batch.

\begin{figure}[t!]
	\centering
	\includegraphics[width=0.95\linewidth]{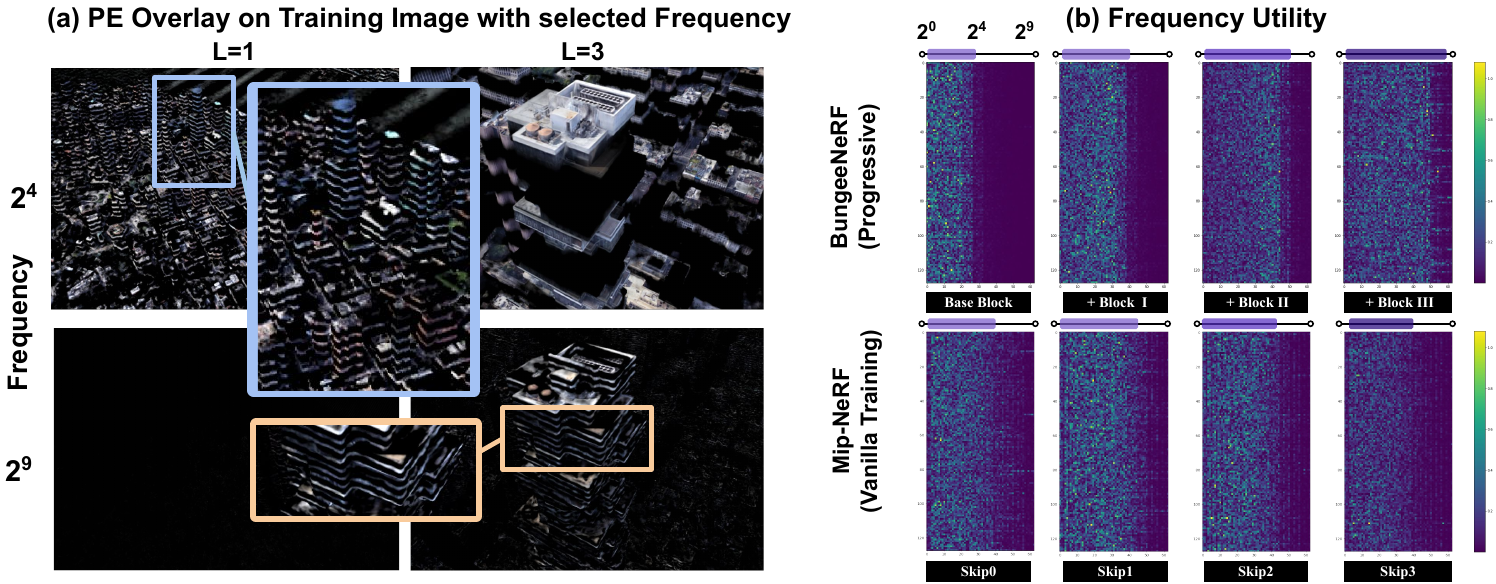}
	\caption{ \small
		(a) Imagery at different scales requires different Fourier feature frequencies in positional encoding to recover corresponding details.
		For example, rendering close views ($L$=3) requires higher-frequency components with $m=10$, while lower-frequency components with $m$=5 are sufficient for remote views ($L$=1).
		(b) \modelname's progressive neural radiance field effectively activates higher-frequency Fourier features in position encoding at deeper blocks, whereas MipNeRF trained under the vanilla scheme is biased to use only lower-frequency Fourier features, even at the deepest skip connection.
	}
	\label{fig:observation_frequency}
\end{figure}

\subsection{Challenges}
\label{subsec:observation}
The challenges brought by drastic multi-scale scene rendering are manifold.
Firstly, the regions observed by close cameras is a subspace of remote cameras, whilst remote cameras captures significant parts of the scene that are not covered by close ones.
This results in inconsistent quality within rendered images, as illustrated in Fig.~\ref{fig:observation_joint_separate}(a).
In contrast, training each scale separately eliminates such inconsistency but sacrifices the communication between successive scales, leading to significant discrepancies as shown in Fig.~\ref{fig:observation_joint_separate}(b), 
which requires further fusion steps to blend the results.

It is noted that
the \emph{effective frequency channels} in PE and IPE differ from one scale to another.
As depicted in Fig.~\ref{fig:observation_frequency}(a)\footnote{The visualizations are acquired by inferring point weights from a trained Mip-NeRF, and accumulate only the selected frequency channel values, following a similar approach of Eq.~\ref{eq:integral} by replacing $\mathbf{c}_{k}$ with the selected channel value for each point.}, 
for a close view ($L$=3) showing complex details of a rooftop,
the low-frequency Fourier feature appears to be insufficient,
while a higher-frequency Fourier feature is activated to better align with such details. 
In contrast, the remote view ($L$=1) can be well represented by the low-frequency Fourier feature, hence the high-frequency one is dampened.
Subsequently, the high-frequency channels in PE are only activated in close views. 
However, due to the limited amount of close views in the training data,  NeRF/MipNeRF trained on all scales as a whole tend to overlook these high-frequency scene components, 
leading to compromised solutions that are biased to utilize low-frequency features only.

\begin{figure*}[t]
	\centering
	\includegraphics[width=\linewidth]{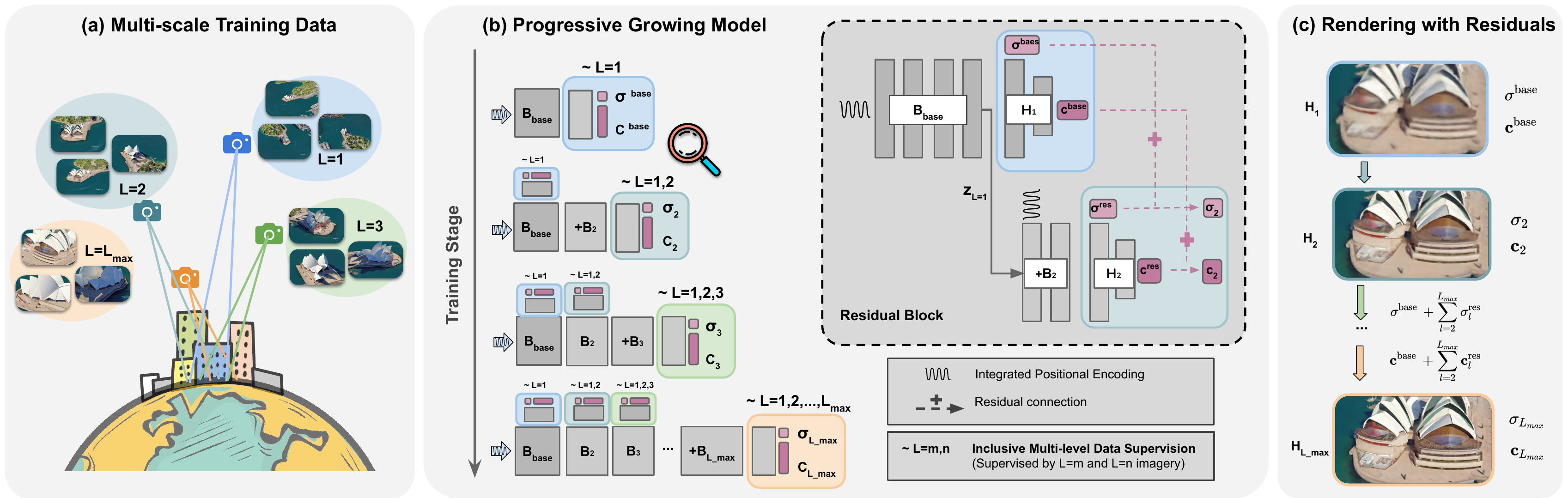}
	\caption{\small Overview of \modelname. 
		(a) An illustration of the multi-scale data in city scenes,
		where we use 
		$L \in \{
		\colorbox{lavender}{1},
		\colorbox{gainsboro}{2},
		\colorbox{gainsboro2}{3}, \ldots
		\}$ to denote scales.
		At each stage, our model grows in synchronization with the training set.
		(b) New residual blocks are appended to the network as the training proceeds, 
		supervised by the union of samples from the most remote scale up to the current scale. The paradigm of a residual block is shown in the dashed box. 
		(c) LOD rendering results obtained at different residual blocks. From shallow to deep, details are added bit by bit.
	(src: \emph{Sydney} scene \copyrightgoogle)}
	\label{fig:training_scheme}
\end{figure*}

\subsection{Progressive Model with Multi-level Supervision}
\label{subsec:method}
The large-scale data change induced by the drastic multi-scale characteristic implies varying learning difficulty and focus. We therefore propose to build and train the model in a progressive manner, with the aim to encourage the division of works among network layers, and unleash the power of the full frequency band in PE.
Moreover, insights from curriculum learning \cite{Guo2018CurriculumNetWS,Zhou2021RobustCL,Guo2020BreakingTC,Dai2020DANASDA,karras2017progressive,Karras2020AnalyzingAI,soviany2021curriculum} tell that 
training models in a meaningful order may ease the training on difficult tasks with model weights initialized on easy samples.
This further underpins our motivation to adopt a progressive training strategy.
The overall paradigm of \modelname is presented in Fig.~\ref{fig:training_scheme}.

In our framework, the training data and the model are grown in a synchronized multi-stage fashion.
We denote $L_{max}$ as the total number of training stages pre-determined for a captured scene, which also discretizes the continuous distance between the camera and the scene.
In experiments, we partition the training data according to the camera \altitude, approximating a hierarchy of resolutions of objects in the scene abide by projective geometry. 
We regard the closest view as at full resolution.
A new scale is set when the camera zooms out by a factor of $\{2^1,2^2,2^3,...\}$.\footnote{In general cases where the distance/depth information are not accessible, $I_l$ can be approximated by the spatial size of textures in the image.
The choice of $L_{max}$ is relatively flexible since it is natural to interpolate results obtained from successive blocks and achieve smooth LOD transition.}
Each image is then assigned with a stage indicator $I_l$ that is shared among all its pixels and the sampling points along casted rays, with $\{I_l=L\}$ denoting the set of images belong to the scale $L$.\footnote{Per-pixel assigned scale is also possible and is likely to gain improvements if depth value is available. For our experiments, image-wise assignment already suffices.}

Start from remote views ($L$=1), as the training progresses, 
views from one closer scale $L+1$ are incorporated at each training stage. 
Such data feeding scheme remedies the bias in sample distribution by allowing the model to put more effort on peripheral regions at the early training stage.
Meanwhile, a rough scene layout can be constructed, which naturally serves as a foundation for closer views in subsequent training stages.
Along with the expansion of the training set, the model grows by appending new blocks. 
As illustrated in Fig.~\ref{fig:training_scheme}, each block is paired with an output head to predict the color and density \emph{residuals} of scene contents viewed at successively closer scales.
The most remote views are allowed to exit from the shallow block with only base color, while close-up views have to be processed by deeper blocks and rendered with progressively added residual colors.
PE is injected to each block via a skip connection to capture the emerging complex details in scene components. 
All layers in the network remain trainable throughout the training process. 

\noindent \textbf{Residual Block Structure.} 
It can be observed that remote views usually exhibit less complex details, making it a relatively easier task to start with. 
We adopt a shallow MLP to be our \emph{base block}, denoted as $B_{base}$, with $D_{base}$=4 hidden layers and $W$=256 hidden units each to fit the most remote scale $\{I_l=1\}$. A skip connection is not included as the base block is shallow.
The output head for color and density follows the original NeRF paper.

When we proceed to the next training stage, a block $B_L$ consisting of $D_{res}$=2 layers of non-linear mappings is appended to the model. A skip connection is added to forward the positional encoding $\gamma(\mathbf{x})$ to the block. 
The intuition is that, since shallow layers are fitted on remote views, the features are learnt to match with low level of detail, hence only low-frequency channels in PE are activated.
However, the new layers need to access the high-frequency channels in PE to construct the emerging details in closer views.
As verified in Fig.~\ref{fig:observation_frequency}(b), our progressive training strategy is able to resort to higher-frequency Fourier features at a deeper block.
In contrast, the matching baseline~(\ie Mip-NeRF-full) is incapable of activating high-frequency channels in PE even after the deepest skip layer, thus failing to represent more complex details.

The additive block $B_L$ outputs residual colors and densities based on the latent features $\mathbf{z}_{L-1}$ obtained from the last mapping layer of the previous block:
\begin{equation}
(\mathbf{c}_{L}^{res}, \sigma_{L}^{res}) = f_{L}^{res}(\mathbf{z}_{L-1}, \mathbf{x}, \mathbf{d}).
\end{equation}
The output exit from head $H_L$ is then aggregated as
\begin{equation}
\mathbf{c}_{L} = \mathbf{c}^{base} + \sum_{l=2}^{L} \mathbf{c}_{l}^{res}, \quad
\sigma_{L} = \sigma^{base} + \sum_{l=2}^{L} \sigma_{l}^{res}.
\end{equation}
The design with residuals has mutual benefits for all scales.
Firstly, it encourages intermediate blocks to concentrate on the missing details and take advance of the high-frequency Fourier features supplied via skip connection.
Furthermore,
it enables gradients obtained from latter blocks to smoothly flow back to earlier blocks and enhance the shallow features with the supervision from closer views.

\noindent \textbf{Inclusive Multi-level Supervision.} To guarantee a consistent rendering quality across all scales, 
at training stage $L$,
the output head $H_L$ is supervised by the union of images from previous scales,~\ie $\{I_l \leq L\}$.
The loss at stage $L$ is aggregated over all previous output heads from $H_1$ to $H_L$: 

\begin{equation}
\begin{aligned}
\mathcal{L}_L = \sum_{l=1}^{L} \sum_{\mathbf{r} \in \mathcal{R}_l}
\left(\left\|\hat{C}(\mathbf{r})-C(\mathbf{r})\right\|_{2}^{2}
\right), 
\end{aligned}
\label{eq:loss_ours}
\end{equation}
where $\mathcal{R}_l$ is the set of rays with stage indicator up to stage $l$, and $C(\mathbf{r}), \hat{C}(\mathbf{r})$ are the ground truth and predicted RGB.

The design of multi-level supervision embeds the idea of \emph{level-of-detail}, 
with deeper output heads providing more complex details in the rendered views.
Compared to traditional mipmapping~\cite{williams1983pyramidal} which requires a pyramid of pre-defined models at each scale, this strategy unifies different levels of detail into a single model and can be controlled with $L$.

%% file: sections/experiment.tex
\section{Experiment}
\label{sec:exp}
We train \modelname on multi-scale city data acquired from Google Earth Studio~\cite{google_earth_studio} and evaluate the quality of reconstructed views and synthesized novel views. 
More experiments are also conducted on landscape scenes, real-world UAV captured scenes, and Blender-synthetic scenes.
We compare our method to NeRF\cite{mildenhall2020nerf}, NeRF with windowed positional encoding (NeRF w/ WPE) \cite{park2021nerfies}, and Mip-NeRF\cite{barron2021mip}.
The effects of the progressive strategy, and the block design with skip layer and residual connection are further analyzed in the ablation study.

\noindent\textbf{Data Collection.} Google Earth Studio~\cite{google_earth_studio} is used as the main data source for our experiments, considering their easy capturing of multi-scale city imagery by specifying camera positions,
with sufficient data quality and rich functionalities to mimic the challenges in real world.
We test our model and compare it against baselines on twelve city scenes across the world. When collecting data, we move the camera in a circular motion and gradually elevate the camera from a low altitude ($\sim$ ground-level) to a high altitude ($\sim$ satellite-level). The radius of the orbit trajectory is expanded during camera ascent to ensure a large enough spatial coverage.
Statistics of the collected scenes are listed in Tab.~\ref{tab:scenes}. Additional data sources are introduced and experimented in Sec.~\ref{subsec:extensions}.

\noindent\textbf{Metrics.} 
We report all metrics on the results exit from the last output head in \modelname.
For quantitative comparison, results are evaluated on low-level full reference metrics, including PSNR and SSIM~\cite{Sitzmann2019SceneRN}. Perceptually, we use LPIPS~\cite{Zhang2018TheUE} which utilizes a pre-trained VGG encoder~\cite{Simonyan2015VeryDC}.
Since the scenes are captured across different scales, a single mean metric averaged from all rendering results cannot fairly reflect quality considering the varied detail complexity across scales. 
We additionally report the mean PSNR obtained at each scale.

\noindent\textbf{Implementation.} We set $L_{max}$=4 for \modelname, hence the model at the final training stage has $10$ layers of $256$ hidden units, with skip connections at the $4,6,8$-th layer. 
For a fair comparison, the best baseline model with the same configuration is also implemented.
Mip-NeRF's IPE \cite{barron2021mip} is adopted in \modelname as we found it consistently outperforms the vanilla PE in multi-scale scenes.
For all methods, the highest frequency is set to $M$=10, with $128$ sample queries per ray. 
We train \modelname for $100k$ iterations per stage and the baselines till converge.
All models are optimized using Adam~\cite{kingma2014adam} with a learning rate decayed exponentially from $5e^{-4}$ and a batch size of $2,048$. The optimizer is reset at each training stage for \modelname.

\input{tables/8scenes}
\input{tables/multiscale_average}

\subsection{Experiment Results} 
\begin{figure*}[t!]
	\centering
	\includegraphics[width=\linewidth]{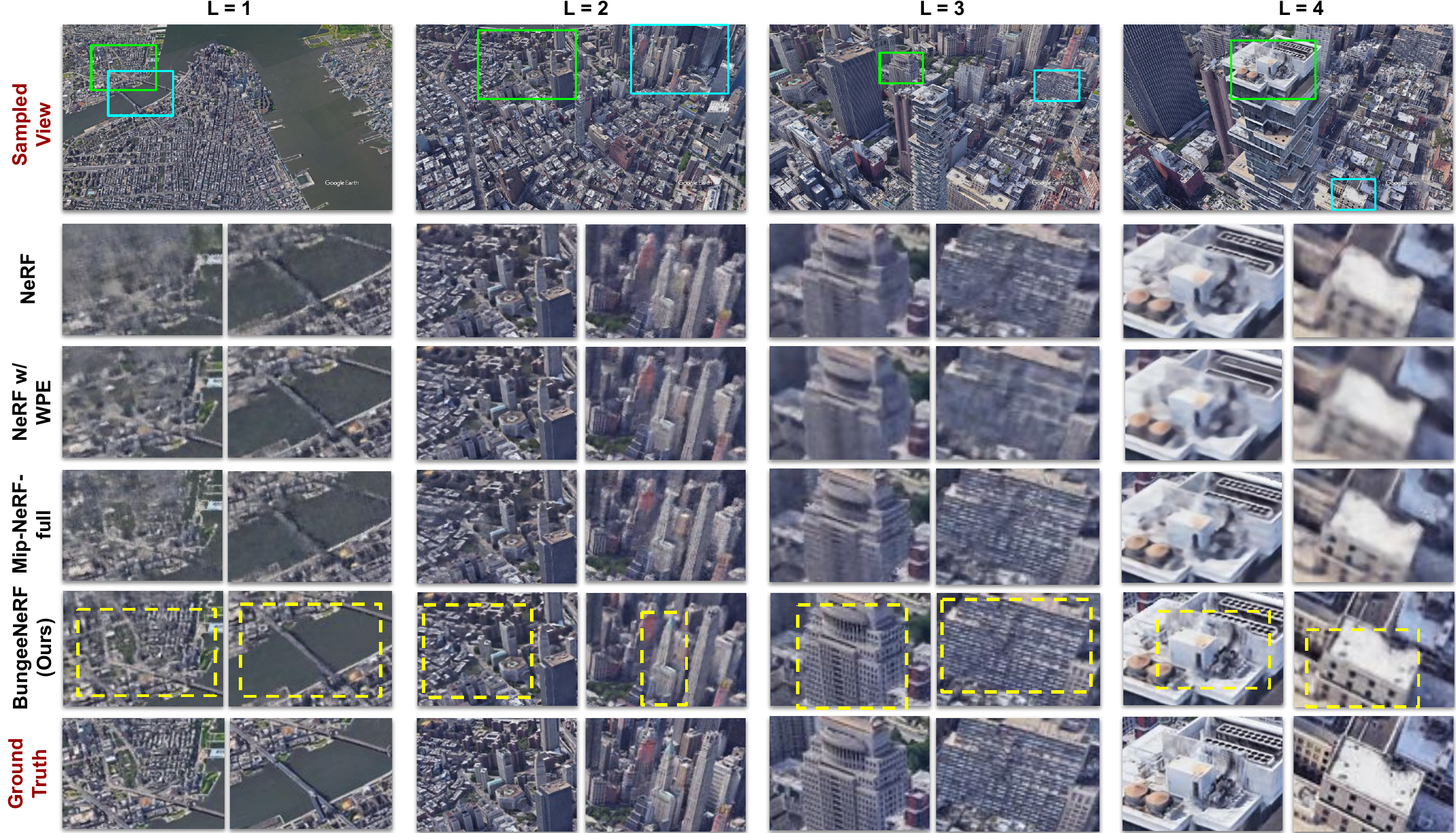}
	\caption{\small Qualitative comparisons between NeRF\cite{mildenhall2020nerf}, NeRF w/ WPE\cite{park2021nerfies}, Mip-NeRF-full\cite{barron2021mip}, and \modelname. \modelname consistently outperforms baseline methods across all scales with reliably constructed details. We strongly encourage readers for more results on diverse scenes in supplementary. (src: \emph{New York} scene \copyrightgoogle)}
	\label{fig:main_exp}
\end{figure*}

Tab.~\ref{tab:main_results} shows our experiment results obtained at the final training phase on two populated city scenes.
Extras scenes listed in Tab.~\ref{tab:scenes} serve as various illustrations throughout the paper, with qualitative results provided in Supplementary.
In general, \modelname achieves PSNR gains of $0.5\sim5$ dB compared to MipNeRF, especially at close scales rich in geometry details.
Fig.~\ref{fig:main_exp} shows the rendered novel views with \modelname and baseline methods on \emph{New York} scene. 

We show that naively deepening the network cannot resolve the problems arisen from different levels of detail and \coverage among scales, where remote views still suffer from blurry artifacts at edges.
On the other hand, \modelname attains a superior visual quality both on the entire dataset and \emph{at each scale} on all metrics,
with a notable improvement in remote scales, where peripheral areas in the rendered views appear to be clearer and more complete compared to jointly training on all images.
When approaching the central target as the camera descends, \modelname continuously brings more details to the scene components, whereas baseline methods always result in blurry background.

Fig.~\ref{fig:observation_frequency}(b) visualizes the network weights associated with each channel of PE, where a clear shift towards the higher frequency portion indicates that the model is able to leverage these parts of the information to construct details in the view. It is also noted that manually truncating PE for different blocks was experimentally found inferior than letting the model learn to choose from all frequencies in our experiments with large-scale changes.

\begin{figure}[t!]
	\centering
	\includegraphics[width=\linewidth]{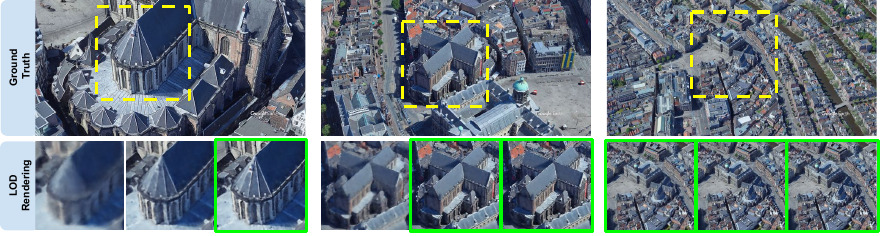}
	\caption{\small Rendering results from different output heads. \modelname allows flexible exits from different residual blocks with controllable LOD. 
		Green box: remote views can exit from earlier heads with sufficient image details, and close views can get finer details when exiting from latter blocks. (src: \emph{Amsterdam} scene \copyrightgoogle)
	}
	\label{fig:lod}
\end{figure}

Fig.~\ref{fig:lod} qualitatively shows the rendering results from different output heads. It can be noticed that $H_2$ produces the coarsest visual results, which omits a significant amount of details when zooming in to closer views, but appears to be plausible for the set of remote views.
The latter output heads gradually add more complex geometric and texture details to the coarse output from previous stages, while maintaining the features learnt at shallower layers meaningful to earlier output heads. In practice, one may consider using earlier heads for rendering remote views for the sake of storage and time consumption.

\subsection{Ablation Study}
\noindent \textbf{Effectiveness of Progressive Strategy.} 
The effectiveness of progressive learning strategy is analyzed through the corresponding three ablation studies: 1) Ablate the inclusive multi-level data supervision,~\ie replacing the output head supervision $\{I_l \leq L\}$ with $\{I_l=L\}$; 2) Ablate the progressive data feeding schedule with fixed model size $L$=4,~\ie trained on all scales simultaneously, with different scales predicted by different output heads. 
3) Ablate model progression and only use the output head $H_4$ of the final block. Note we still adopt the progressive data feeding schedule for this case.
Results are listed in Tab.~\ref{tab:ablate_results}.

\input{tables/ablate}

It is observed that: 1) Without the inclusive multi-level data supervision, 
the model achieves a slightly higher PSNR at the closest scale ($L$=4), but the performances at remote scales degrade drastically. 
This is because deeper blocks are solely trained on close views, and are not responsible for constructing remote views.
As the result, the additive blocks might deviate to better fit the close views, whilst shallow layers just maintain their status and ignore the extra information from deeper layers. 
2) Without the progressive data feeding strategy, the results are even inferior than baselines. 
Firstly, due to the residual connection, the suboptimal results from remote views set a less ideal foundation for the modeling of closer views.
Secondly, with all data being fitted simultaneously, the effective channels in PE still can not be distinguished between scales. On top of this, regularizing shallow features with remote views puts more restrictions on model capacity, which further harms the performance.  
3) Without appending new layers, it is difficult for the model to accommodate the newly involved high-frequency information from closer scales, 
where the model has already been well fitted on distant scales. As the result, it achieves decent performance on the most remote scale ($L$=1) but becomes worse at closer scales. 

\noindent \textbf{Effectiveness of Model Design.}
To determine the effectiveness of the residual connection and the skip connection in our newly introduced block structure for neural radiance fields, we run ablations on: 1) Discarding the residual connection when growing the network; 2) Inserting a skip connection at the same position as the original NeRF and discarding the rest in the additive blocks.

\begin{figure}[t!]
	\centering
	\includegraphics[width=\linewidth]{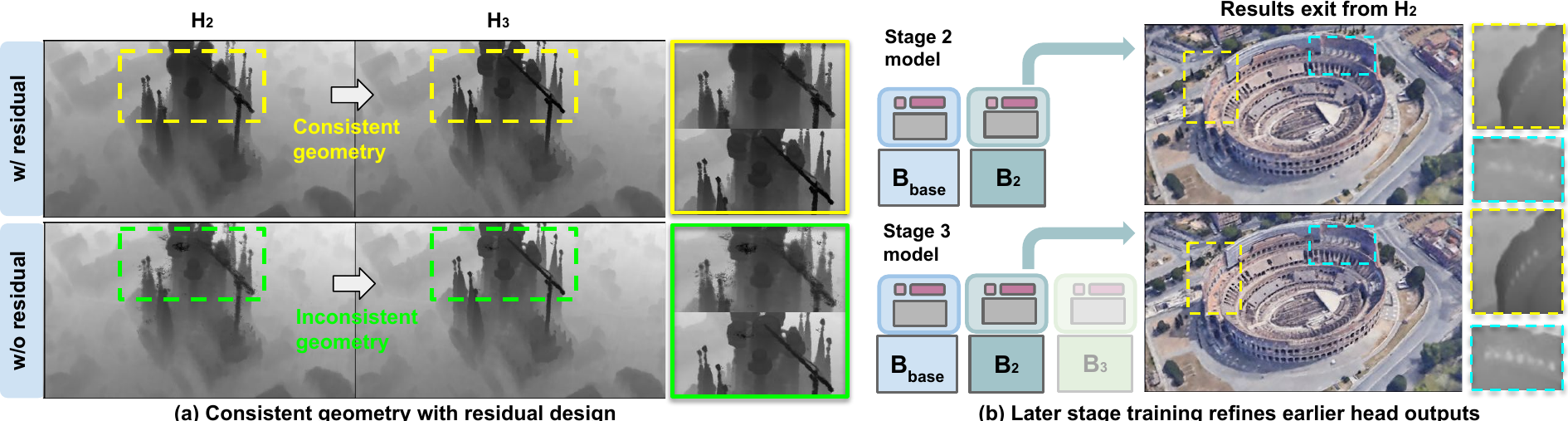}
	\caption{\small The residual connection enables supervisions from latter blocks to help refine the outputs from earlier heads (Fig.(b)), yielding more accurate geometries with consistent depths across scales (Fig.(a)). It also helps \modelname to focus on the missing details between the results rendered by shallower blocks and ground truth, leading to sharper visuals. 
		(src: \emph{Barcelona} and \emph{Rome} scenes \copyrightgoogle)
	}
	\label{fig:residual}
\end{figure}

Tab.~\ref{tab:ablate_results} shows that, without the skip layer or the residual connection, performance at each scale slightly degenerates but still outperforms baselines. 
1) It can be observed that, the influence of skip layer is more notable on closer scales.
In particular, on \emph{Transamerica} scene, the absence of skip connection severely harms the performance at scale $L$=4, 
which could be ascribed to the fact that its remote views are much easier than close views. 
2) Residual connections help form a better scene feature at remote scales for shallow output heads, by leveraging the supervision from deeper blocks to help correct the errors in early training phases, as shown in Fig.~\ref{fig:residual}. Consequently, rendering results of lower LOD are more accurate, which in turn benefit latter training phases on closer views. Furthermore, we found that as the training proceeds, \modelname with residual gains an increasing advantage compared to the counterpart without residual. The additive nature of residual densities and colors enforces the model to emphasize the errors between the images rendered by previous stages and the ground truth, guiding the deeper blocks to recover the missing details using the newly introduced frequency information. Meanwhile, it also improves the predictions from previous output heads, yielding better base geometries and colors.

%% file: tables/8scenes.tex
\begin{table*}[t]
	\centering
	\resizebox{\linewidth}{!}{
		\begin{tabular}{l|cc|cccccccccc} 
			& \multicolumn{2}{c|}{Main Experiments} & \multicolumn{10}{c}{Extra Tests: Used for various illustrations throughout the paper and additional comparisons in Supplementary}  \\
			City Scene & New York &  San Francisco & Sydney & Seattle & Chicago & Quebec & Amsterdam & Barcelona & Rome  & Los Angeles & Bilbao & Paris \\
			
			Alt (m) & 56 Leonard &  Transamerica & Opera House & Space Needle & Pritzker Pavilion & Château Frontenac & New Church & Sagrada Familia & Colosseum  & Hollywood & Guggenheim & Pompidou \\ \hline
			
			Lowest &                    290 &                    326 &                    115 &                    271 &                    365 &                    166 &                    95 &                    299 &                    130 &                    660 &                   163 &                    159   \\
			Highest &                    3,389 &                    2,962 &                    2,136 &                    18,091 &                    6,511 &                    3,390 &                    2,3509 &                    8,524 &                    8,225 &                    12,642 &                    7,260 &                    2,710               \\
			\hline
			\# Images &  463 &  455 &  \multicolumn{10}{c}{220} \\
		\end{tabular}}
	\caption{\small Twelve city scenes captured in Google Earth Studio.}
	\label{tab:scenes}
\end{table*}

%% file: tables/multiscale_average.tex
\begin{table*}[t]
	\centering
	\resizebox{\linewidth}{!}{
		\begin{tabular}{l|cccc|cccc|ccc|ccc} 
			& \multicolumn{4}{c|}{\emph{56 Leonard} (PSNR $\uparrow$)} & \multicolumn{4}{c|}{ \emph{Transamerica} (PSNR $\uparrow$)} & \multicolumn{3}{c|}{\emph{56 Leonard} (Avg.)} & \multicolumn{3}{c}{\emph{Transamerica} (Avg.)} \\
			& Stage I. & Stage II. & Stage III. & Stage IV. & Stage I. & Stage II. & Stage III. & Stage IV. & PSNR $\uparrow$ & LPIPS $\downarrow$ & SSIM $\uparrow$ & PSNR $\uparrow$ & LPIPS $\downarrow$ & SSIM $\uparrow$ \\ \hline
			
			NeRF (D=8, Skip=4)&                    21.279 &                    22.053 &                    22.100 &                    21.853 &                    22.711 &                    22.811 &      \underline{22.976} &    \underline{21.581} &                    21.702 &                    0.320 &                    0.636 &                    22.642 &                    0.318 &  0.690 \\
			NeRF w/ WPE (D=8, Skip=4)&                    22.022 &                    22.097 &                    21.799 &                    21.439 &            \underline{23.382} &     \underline{23.171} &                    22.450 &                    20.806 &                    21.672 &                    0.365 &                    0.633 &                    22.352 &                    0.331 & 0.680\\
			Mip-NeRF-small (D=8, Skip=4)                                                 &                    21.899 &                    22.179 &                    22.045 &                    21.763 &                    23.346 &                    23.030 &                    22.645 &                    20.937 &                    21.975 &                    0.344 &                    0.648 &                    22.692 &                    0.327  & 0.687\\
			Mip-NeRF-large (D=10, Skip=4)                                                   &                    22.020 &                    22.403 &                    22.277 &                    21.975 &                    23.196 &                    22.900 &                    22.439 &                    20.743 &                    22.227 &                    0.318 &                    0.666 &                    22.525 &                    0.330  & 0.686\\

			 Mip-NeRF-full (D=10, Skip=4,6,8)                                                   &       \underline{22.043} & \underline{22.484} &       \underline{22.690} & \underline{22.355} &      23.377 &     23.156 &      22.854 &        21.152 & \underline{22.312} &  \underline{0.266} & \underline{0.689} & \underline{22.828} &\underline{0.314}  &\underline{0.699}\\
\hline
\modelname (same iter as baselines) &  \bf{23.145} &  \bf{23.548} &  \bf{23.744} & \bf{23.015} & \bf{24.259} &  \bf{23.911} &  \bf{23.507} &  \bf{22.942} &  \bf{23.481} &  \bf{0.235} & \bf{0.739} & \bf{23.606} &    \bf{0.265}  &  \bf{0.749}\\
\modelname (until convergence)                                  &   \bf{24.120} &  \bf{24.345} &  \bf{25.382} &   \bf{25.112} &   \bf{24.608} &  \bf{24.350} &  \bf{24.357} &  \bf{24.608} &    \bf{24.513} &    \bf{0.160} & \bf{0.815} &    \bf{24.415} & \bf{0.192}  & \bf{0.801} \\
		\end{tabular}
	}
	\caption{\small Quantitative comparison on \emph{56 Leonard} and \emph{Transamerica} scenes. $D$ denotes model depth and $Skip$ indicates which layer(s) the skip connection is inserted to. Better performance can be achieved if each stage is trained until convergence. (\textbf{best}/\underline{2nd best})}
	\label{tab:main_results}
\end{table*}

%% file: tables/ablate.tex
\begin{table*}[t!]
	\centering
	\resizebox{\linewidth}{!}{
		\begin{tabular}{l|cccc|cccc|ccc|ccc}
			& \multicolumn{4}{c|}{\emph{56 Leonard} (PSNR $\uparrow$)} & \multicolumn{4}{c|}{ \emph{Transamerica} (PSNR $\uparrow$)} & \multicolumn{3}{c|}{\emph{56 Leonard} (Avg.)} & \multicolumn{3}{c}{\emph{Transamerica} (Avg.)} \\
			& Stage I. & Stage II. & Stage III. & Stage IV. & Stage I. & Stage II. & Stage III. & Stage IV. & PSNR $\uparrow$ & LPIPS $\downarrow$ & SSIM $\uparrow$ & PSNR $\uparrow$ & LPIPS $\downarrow$ & SSIM $\uparrow$ \\ \hline
			\bf{Full \modelname} &  \bf{23.145} &  \bf{23.548} &  \bf{23.744} & \bf{23.015} & \bf{24.259} &  \bf{23.911} &  \bf{23.507} &  \bf{22.942} &  \bf{23.481} &  \bf{0.235} & \bf{0.739} & \bf{23.606} &    \bf{0.265}  &  \bf{0.749}\\ \hline
			- inclusive supervision                                                 &                    22.570 &                    22.475&                    23.300 &                    23.188 &      23.623               &     23.039                &       22.508              &       21.294              &                    22.773 &                    0.298 &                    0.700 &           22.789          &          0.309            & 0.696 \\
			- data feeding scheme                                                &   22.268                    &       22.660              &      22.380               &  21.949                  &       23.563              &            23.355         &             23.015        &      21.366               &           22.340          &    0.324                 &    0.679                 &      23.015               &     0.310                 & 0.711 \\
			- model growing                                            &   22.826 &   22.121 &   21.471 &   20.972 &  24.026     &  23.560    &  22.554    &  20.503       &  22.139  & 0.326    & 0.666    &  23.055   & 0.317    & 0.706 \\
			\hline
			- skip layer &    23.046 &                 23.364 &                    23.317 &                    22.746 &                    24.232 &                    23.696 &                    22.950 &                    22.151 &                    23.159 &                    0.259 &                   0.730 &                    23.235 &                    0.283 &  0.725 \\
- residual&     23.104 &    23.325 &                   23.228 &                   22.786 &                    24.091 &                    23.608 &                    23.236 &                    22.796 &                    23.139 &                    0.257 &                   0.729 &                    23.364 &                     0.272 & 0.735\\ 
		\end{tabular}
	}
	\caption{\small Ablation studies on \emph{56 Leonard} and \emph{Transamerica Pyramid} scene. The first set ablates the progressive strategy and the second set ablates residual block designs.}
	\label{tab:ablate_results}
\end{table*}

%% file: sections/ablation.tex
\subsection{Extensions}
\label{subsec:extensions}
\begin{figure}[t!]
	\centering
	\includegraphics[width=\linewidth]{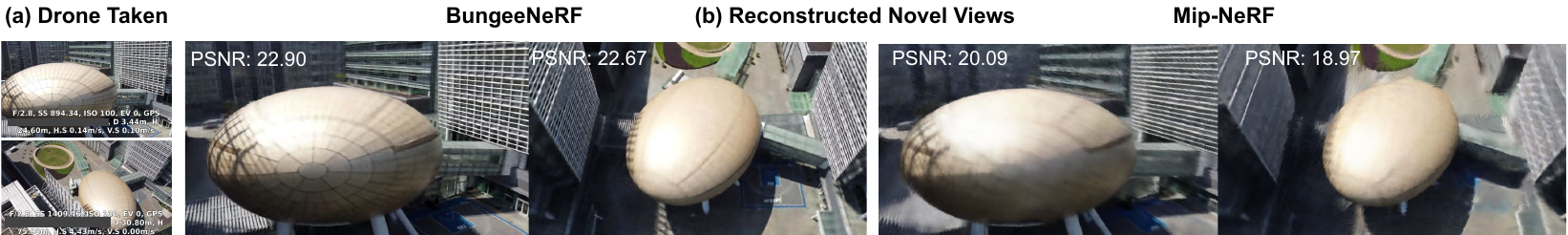}
	\caption{\small Results on UAV captured scenes.
		(a) Recorded camera information shows the wide range of camera altitude changes (24$\sim$76m). (b) Novel views rendered with \modelname (a simpilied $L$=2 setting) demonstrate the superior performance compared to Mip-NeRF, which suggests its practical usage in real-world applications.}
	\label{fig:drone}
\end{figure}

\begin{figure}[t!]
	\centering
	\includegraphics[width=\linewidth]{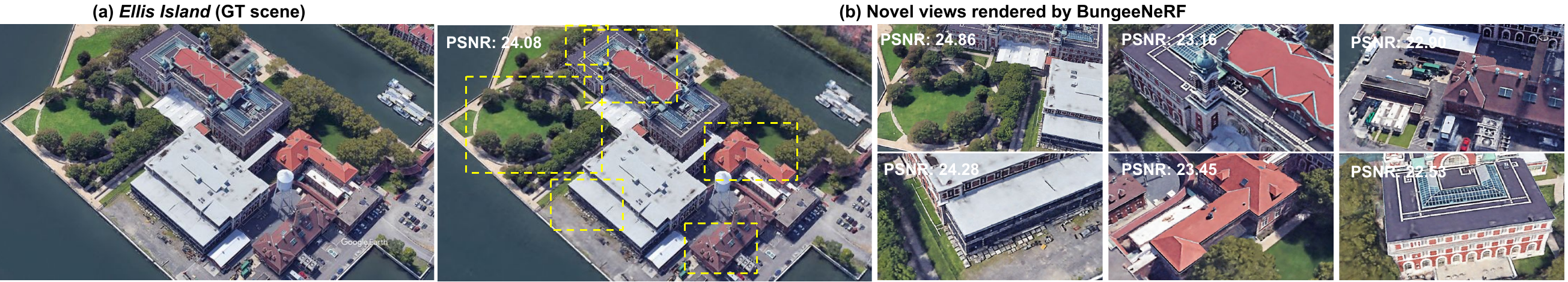}
	\caption{\small \modelname (trained with $L$=3 stages) successfully recovers the fine details in a multi-dive city scene. (src: \emph{Ellis Island, NY}\copyrightgoogle)}
	\label{fig:ellis}
\end{figure}

Apart from city scenes, experiments are also conducted on landscape (in supplementary) and Blender-synthetic scenes (Fig.~\ref{fig:blender}). 
\modelname can also represent scale changes expanding earth-level scenes as shown in Fig.~\ref{fig:teaser}.
Additional to single dive-in and -out camera trajectory, Fig.~\ref{fig:ellis} shows results obtained on a scene with multi-dive flying pattern, where \modelname effectively recovers details of multiple targets.
Besides synthetic scenes, we also tested on UAV captured real-world scene where camera poses are estimated by COLMAP~\cite{schonberger2016structure} (Fig.~\ref{fig:drone}).

\begin{figure}[t!]
	\centering
	\includegraphics[width=\linewidth]{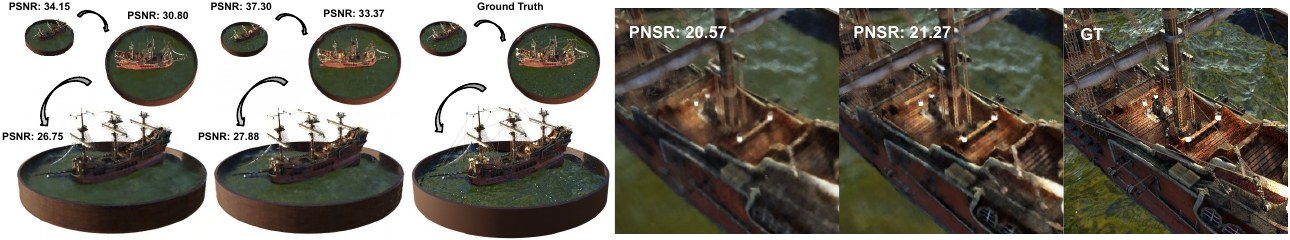}
	\caption{\small Results on Blender Synthetic data showing four scales. From left to right are Mip-NeRF, \modelname (trained with $L$=4 stages), and ground truth images.}
	\label{fig:blender}
\end{figure}

%% file: sections/conclusion.tex
\section{Discussion and Conclusion}
\label{sec:conclusion}
In this work, we propose \modelname, a progressive neural radiance field, to model scenes under a drastic multi-scale setting, with large-scale variation in level of detail and linear field of view,
where a NeRF/Mip-NeRF trained under normal scheme has difficulty in accommodating such extreme data change. 
\modelname adopts a novel progressive training paradigm that synchronously grows the model and training set to learn a hierarchy of scene representations from coarse to fine, which demonstrates superior results on various scenes compared to baselines with ensured high-quality rendering across all scales.

While \modelname functions as a good building block for modeling large-scale 3D scenes in real world which is naturally rich in multi-scale observations, it is natural to consider its combined use with orthogonal advanced neural rendering techniques to bring more high-quality renderings results. Facing such needs,  a comprehensive neural rendering system with integrated merits on multiple characteristics (\eg, large-scale, photorealistic, dynamic, editable, etc) with accurate control could be a promising and interesting research direction in the future.

\noindent\textbf{Acknowledgment}
This work is supported by GRF 14205719, TRS T41-603/20-R, Centre for Perceptual and Interactive Intelligence, and CUHK Interdisciplinary AI Research Institute; and by the ERC Consolidator Grant 4DRepLy (770784).